\def\year{2017}\relax
\documentclass[letterpaper]{article} 
\usepackage{aaai17}  
\usepackage{times}  
\usepackage{helvet}  
\usepackage{courier}  
\usepackage{url}  
\usepackage{graphicx}  
\usepackage[ruled]{algorithm2e}
\usepackage{amsmath}
\usepackage{multirow}
\usepackage{caption}
\usepackage{subcaption}
\begin{document}
%
\title{\textsc{Why}: Natural Explanations from a Robot Navigator}
\author{Raj Korpan\textsuperscript{1}, Susan L. Epstein\textsuperscript{1,2}, Anoop Aroor\textsuperscript{1}, \and Gil Dekel\textsuperscript{2}\\
\textsuperscript{1}The Graduate Center and \textsuperscript{2}Hunter College, City University of New York\\
rkorpan@gradcenter.cuny.edu, susan.epstein@hunter.cuny.edu, aaroor@gradcenter.cuny.edu, gil.dekel47@myhunter.cuny.edu\\
}

\maketitle
\begin{abstract}
Effective collaboration between a robot and a person requires natural communication. When a robot travels with a human companion, the robot should be able to explain its navigation behavior in natural language. This paper explains how a cognitively-based, autonomous robot navigation system produces informative, intuitive explanations for its decisions. Language generation here is based upon the robot's commonsense, its qualitative reasoning, and its learned spatial model. This approach produces natural explanations in real time for a robot as it navigates in a large, complex indoor environment.
\end{abstract}

\section{Introduction}
\noindent Successful human-robot collaboration requires \textit{natural explanations}, human-friendly descriptions of the robot's reasoning in natural language. In \textit{collaborative navigation}, a person and an autonomous robot travel together to some destination. The thesis of this paper is that natural explanations for collaborative navigation emerge when a \emph{robot controller} (autonomous navigation system) is cognitively based. This paper introduces \textsc{Why}, an approach that accesses and conveys the robot's reasoning to provide its human companion with insight into its behavior. The principal results presented here are natural explanations from an indoor robot navigator.

Even in unfamiliar, complex spatial environments (\textit{worlds}), people travel without a map to reach their goals successfully~\cite{conlin2009getting}. Efficient human navigators reason over a mental model that incorporates commonsense, spatial knowledge, and multiple heuristics~\cite{golledge1999human}. They then use the same model to explain their chosen path and their reasons for decisions along the way. Our research goal is an autonomous robot navigator that communicates with its human companions much the way people do.

\textsc{Why} explains a navigation decision in natural language. It anticipates three likely questions from a human companion: ``Why did you decide to do that?'' ``Why not do something else?'' and ``How sure are you that this is the right decision?'' \textsc{Why} generates its answers with \textit{SemaFORR}, a robot controller that learns a spatial model from sensor data as it travels through a partially-observable world without a map~\cite{epstein2015learning}. SemaFORR's cognitively-based reasoning and spatial model facilitate natural explanations.

\textsc{Why} is an interpreter; it uses SemaFORR's cognitive foundation to bridge the perceptual and representational gap between human and robot navigators. \textsc{Why} and SemaFORR could accompany any robot controller to provide natural explanations. More broadly, \textsc{Why} can be readily adapted to explain decisions for other applications of FORR, SemaFORR's underlying cognitive architecture.

The next section of this paper reviews related work. Subsequent sections describe SemaFORR and formalize \textsc{Why}. Finally, we evaluate \textsc{Why}-generated explanations and give examples of them as our mobile robot navigates through a large, complex, indoor world.

\section{Related Work}
When a robot represents and reasons about space similarly to the way people do, it facilitates human-robot collaboration~\cite{kennedy2007spatial}.
Communication with a robot allows people to build a mental model of how it perceives and reasons, and thereby helps to establish trust~\cite{kulesza2013too,bussone2015role}. A recent approach grounded perceived objects between the robot and a person to build a mutual mental model, and then generated natural language descriptions from it~\cite{chai2016collaborative}. Although that supported natural dialogue, it did not explain the reasoning that produced the robot's behavior. 

Despite much work on how a robot might understand natural language from a human navigator~\cite{boularias2016learning,duvallet2016inferring,thomason2015learning}, natural explanations from a robot navigator to a person remain an important open problem. Such work has thus far required detailed logs of the robot's experience, which only trained researchers could understand~\cite{landsiedel2017review,scalise2017natural}. It is unreasonable, however, to expect people to decipher robot logs.

Natural language descriptions of a robot's travelled path have addressed abstraction, specificity, and locality~\cite{rosenthal2016verbalization,perera2016dynamic}. A similar approach generated path descriptions to improve sentence correctness, completeness, and conciseness~\cite{barrett2017driving}. Those approaches, however, used a labeled map to generate descriptions and did not explain the robot's reasoning. Other work visually interpreted natural-language navigation commands with a semantic map that showed the robot's resulting action~\cite{oh2016integrated}. Although a person might eventually unpack the robot's reasoning process this way, no natural language explanation was provided.

Researchers have generated navigation instructions in natural language from metric, topological, and semantic information about the world~\cite{daniele16} or rules extracted from human-generated instructions~\cite{dale2005using}. Other work has focused on human spatial cognition~\cite{look2008cognitively}, or on simplicity and understandability~\cite{richter2008simplest}. None of these approaches, however, can explain how the instructions were generated, nor can they justify a particular instruction.

More generally, researchers have sought human-friendly explanations for systems that learn. Trust in and understanding of a learning system improved when people received an explanation of why a system behaved one way and not another~\cite{lim2009and}. Several approaches to sequential tasks have explained Markov decision processes, but the resultant language was not human-friendly and was not based on human reasoning~\cite{ramakrishnan2016towards,dodson2013english,khan2011automatically}. In summary, although intelligent systems should be able to provide natural explanations during collaborative navigation, to the best of our knowledge no work has focused on explanations for the robot's decisions. \textsc{Why} addresses that gap.

\begin{figure}[b]
\centering
\begin{subfigure}[]{.33\linewidth}
    \centering
    \includegraphics[width = \linewidth]{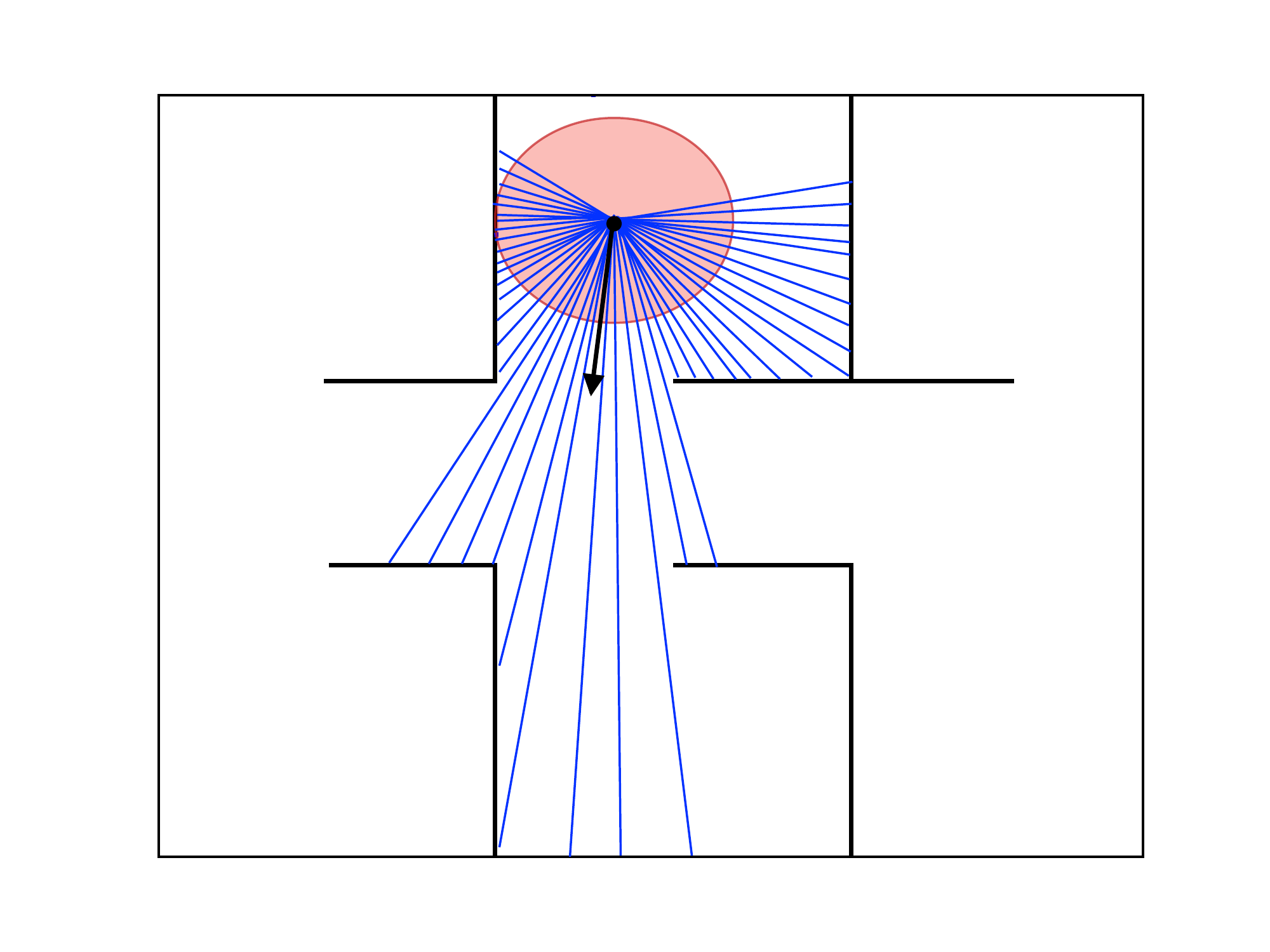}
    \caption{}
\end{subfigure}
\hfill
\begin{subfigure}[]{.33\linewidth}
    \centering
    \includegraphics[width = \linewidth]{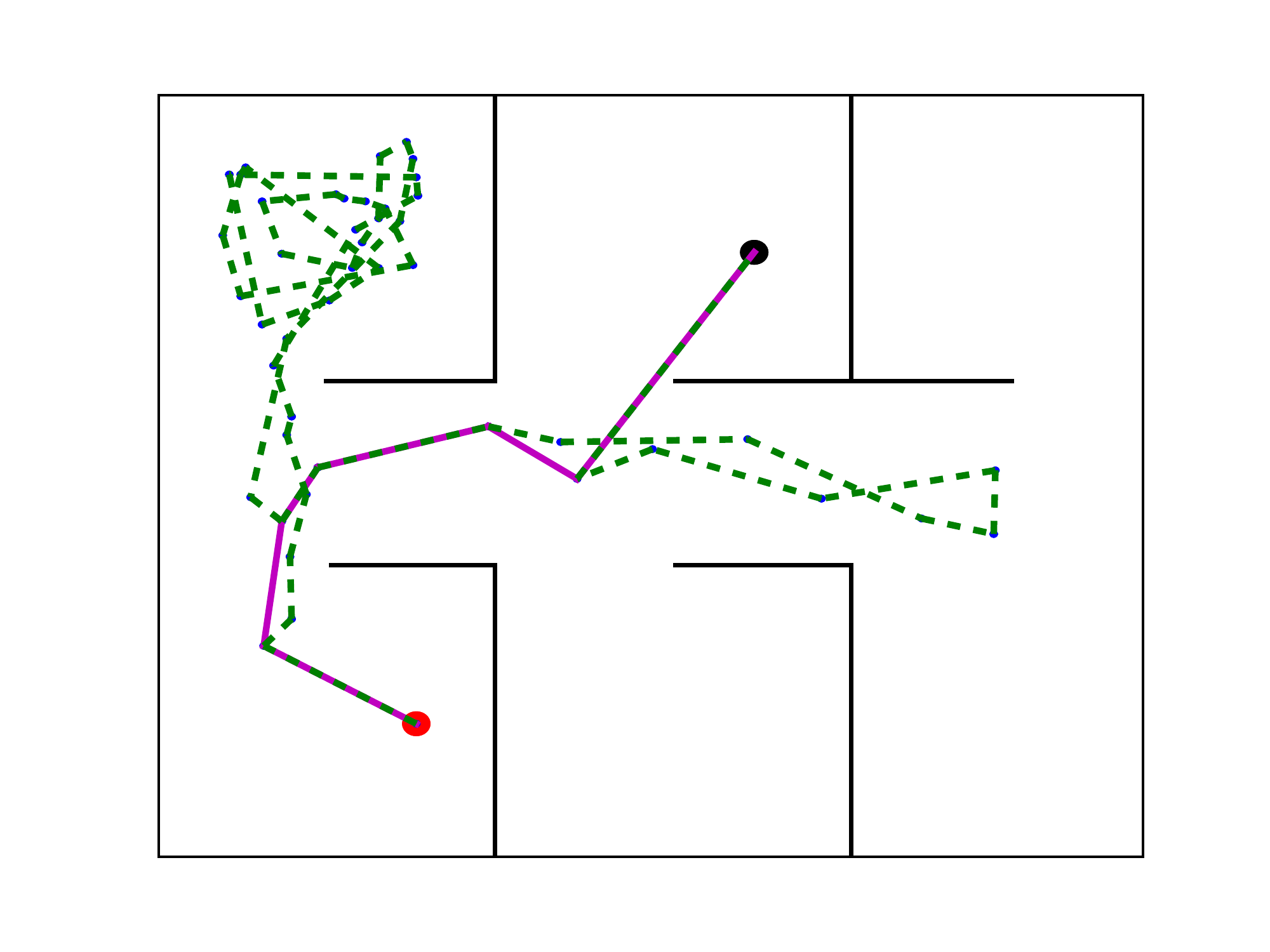}
    \caption{}
\end{subfigure}
\hfill
\begin{subfigure}[]{.32\linewidth}
    \centering
    \includegraphics[width = \linewidth]{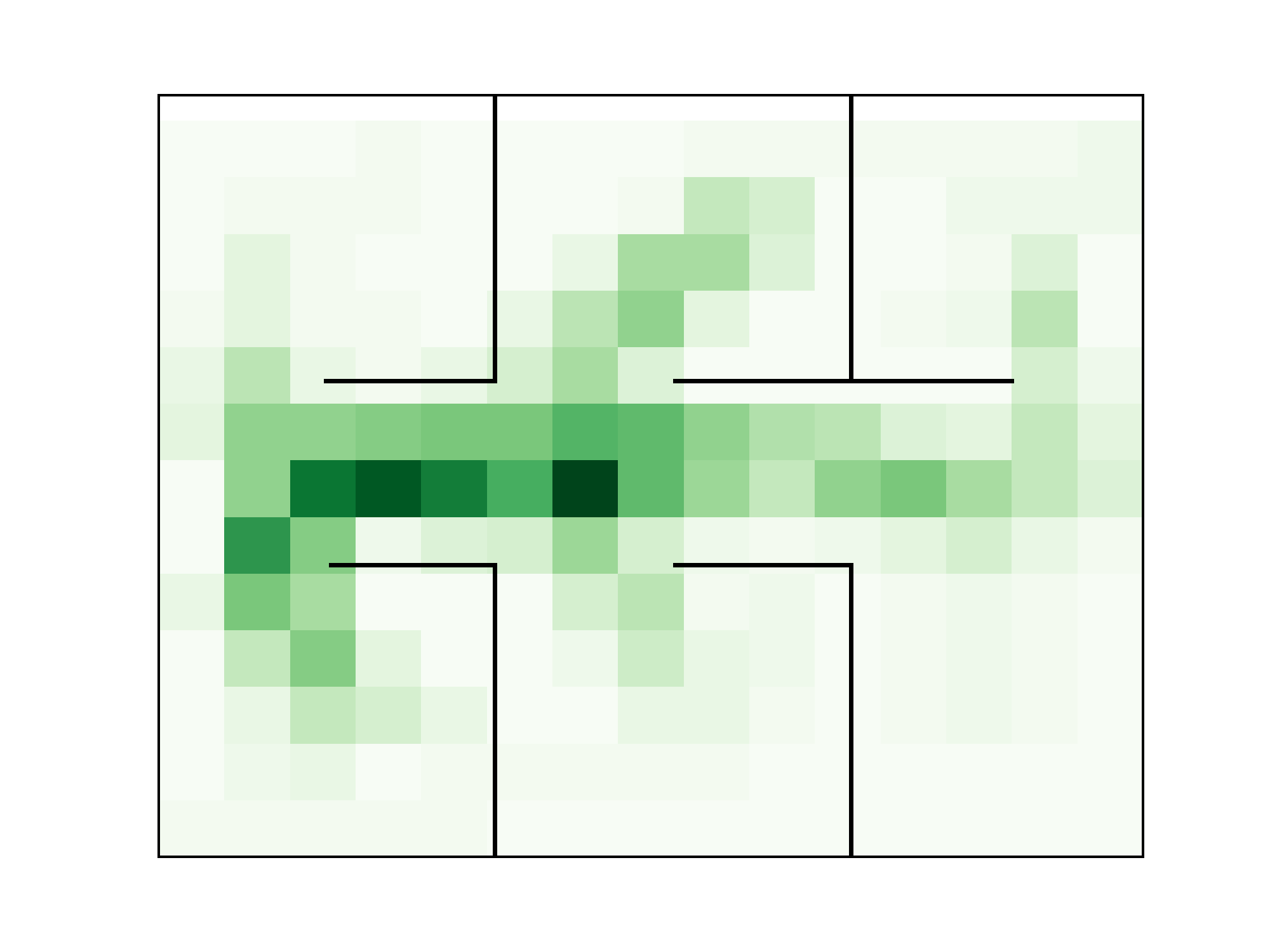}
    \caption{}
\end{subfigure}
\caption{Examples of affordances in a simple world (a) a region (b) a path and its trail (c) conveyors} 
    \label{fig:spatialmodel}
\end{figure}

\section{SemaFORR}
SemaFORR is a robot controller implemented in \textit{ROS}, the state-of-the-art Robot Operating System. SemaFORR selects one action at a time to move the robot to its target location. Instead of a world map, SemaFORR uses local sensor data, learned knowledge, and reactive, heuristic reasoning to contend with any obstacles and reach its target. The resultant behavior is satisficing and human-like rather than optimal.

A \textit{decision state} records the robot's current sensor data and its \textit{pose} $\langle x, y, \theta \rangle$, where $\langle x, y \rangle$ is its location and $\theta$ is its orientation with respect to an allocentric, two-dimensional coordinate system. As the robot travels, its \textit{path} to a target is recorded as a finite sequence of decision states. SemaFORR makes decisions based on a hierarchical reasoning framework and a spatial model that it learns while it navigates. \textsc{Why} uses them both to generate its explanations.

\subsection{Spatial Model}
SemaFORR learns its compact, approximate spatial model from experience. The model captures many of the features of a cognitive map, the representation that people construct as they navigate~\cite{foo2005humans}. Instead of a metric map, SemaFORR's model is a set of \textit{spatial affordances}, abstract representations that preserve salient details and facilitate movement. As the robot travels or once it reaches its target, it learns spatial affordances from local sensor readings and stores them as episodic memory. Figure \ref{fig:spatialmodel} gives examples. 

A \textit{region} is an unobstructed area where the robot can move freely, represented as a circle. A region's center is the robot's location in a decision state; its radius is the minimum distance sensed from the center to any obstacle. An \textit{exit} is a point that affords access to and from a region, learned as a point where the robot crossed the region's circumference . 

A \textit{trail} refines a path the robot has taken. It is an ordered list of \textit{trail markers}, decision states selected from the robot's path. The first and last trail markers are the initial and final decision states on the path. Trail learning works backward from the end of the path; it creates a new trail marker for the earliest decision state that could have sensed the current trail marker. The resultant trail is usually shorter than the original path and provides a more direct route to the target.

A \textit{conveyor} is a small area that facilitates travel. It is represented in a grid superimposed on the world, where each cell tallies the frequency with which trails pass through it. High-count cells in the grid are conveyors. 

The spatial model combines affordances to produce more powerful representations. For example, a \textit{door} generalizes over the exits of a region. It is represented as an arc along the region's circumference. The door-learning algorithm introduces a door when the length of the arc between two exits is within some small $\epsilon$. Once generated, a door incorporates additional exits if they are within $\epsilon$ of it. Another example is the \textit{skeleton}, a graph that captures global connectivity with a node for each region. An edge in the skeleton joins two nodes if a path has ever moved between their corresponding regions. Along with commonsense qualitative reasoning, affordances are used to select the robot's next action.

\subsection{Reasoning Framework}
SemaFORR is an application of \textit{FORR}, a cognitive architecture for learning and problem solving~\cite{epstein1994right}. FORR is both reactive and deliberative. Reactivity supports flexibility and robustness, and is similar to how people experience and move through space~\cite{spiers2008dynamic}. Deliberation makes plans that capitalize on the robot's experience; it is the focus of current work\nocite{aroor2017toward} (Aroor and Epstein, in press).

The crux of any FORR-based system is that good decisions in complex domains are best made reactively, by a mixture of good reasons. FORR represents each good reason by a procedure called an \textit{Advisor}. Given a decision state and a discrete set of possible actions, an Advisor expresses its opinions on possible actions as \textit{comments}. In a \textit{decision cycle}, SemaFORR uses those comments to select an action. Possible actions are alternately a set of forward moves of various lengths or a set of turns in place of various rotations. A move with distance 0 is equivalent to a pause. Thus, in any given decision state, SemaFORR chooses only the intensity level of its next move or turn. The resultant action sequence is expected to move the robot to its target.

\begin{table}[!b]
    \fontsize{9pt}{10pt}\selectfont
    \centering
    \begin{tabular}{p{0.9in} p{2.08in}}
    \hline
    \multicolumn{2}{l}{Tier 1, in order} \\
    \hline
        \textsc{Victory} & Go toward an unobstructed target \\
        \textsc{AvoidWalls} & Do not go within $\epsilon$ of an obstacle \\
        \textsc{NotOpposite} & Do not return to the last orientation \\
    \hline
    \hline
    \multicolumn{2}{l}{Tier 3} \\
    \hline
        \multicolumn{2}{c}{\textit{Based on commonsense reasoning}} \\
        \textsc{BigStep} & Take a long step \\
        \textsc{ElbowRoom} & Get far away from obstacles \\
        \textsc{Explorer} & Go to unfamiliar locations \\
        \textsc{GoAround} & Turn away from nearby obstacles \\
        \textsc{Greedy} & Get close to the target \\
    \hline
        \multicolumn{2}{c}{\textit{Based on the spatial model}} \\
        \textsc{Access} & Go to a region with many doors \\
        \textsc{Convey} & Go to frequent, distant conveyors \\
        \textsc{Enter} & Go into the target's region \\
        \textsc{Exit} & Leave a region without the target \\
        \textsc{Trailer} & Use a trail segment to approach the target \\
        \textsc{Unlikely} & Avoid dead-end regions \\
    \hline
    \end{tabular}
    \caption{SemaFORR's Advisors and their rationales. Tier 2 is outside the scope of this paper.}
    \label{tab:advisors}
\end{table}

\begin{algorithm}[b]
    \fontsize{9pt}{10pt}\selectfont
    \DontPrintSemicolon
    \textbf{Input:} \textit{decision state, target location, spatial model}\;
    \textbf{Output:} \textit{explanation}\;
    \Switch{mode(decision)}{
      \uCase{tier 1 decides action}{
        \textit{explanation} $\leftarrow$ sentence based on \textsc{Victory}\;
      }
      \uCase{only 1 unvetoed action remains after tier 1}{
        \textit{explanation} $\leftarrow$ sentence based on vetoes\;
      }
      \Other{
        Compute \textit{t}-statistics for tier-3 Advisors' strengths\;
        Categorize the support level for the chosen action\;
        Complete template for each Advisor with its support level and rationale\;
        \textit{explanation} $\leftarrow$ combined completed templates\;
      }
    }
    \Return{explanation}
    \caption{\textsc{Why}'s Explanation Procedure}
    \label{alg:SEP}
\end{algorithm}

SemaFORR's Advisors are organized into a three-tier hierarchy, with rules in tier 1 and commonsense, qualitative heuristics in tier 3. Tier 1 invokes its Advisors in a predetermined order; each of them can either mandate or veto an action. If no action is mandated, the remaining, unvetoed actions are forwarded to tier 3. (Natural explanations for tier 2, SemaFORR's deliberative layer, are a focus of current work.) Table \ref{tab:advisors} lists the Advisors' rationales by tier. 

Each tier-3 Advisor constructs its comments on the remaining possible actions with its own commonsense rationale. Comments assign a \textit{strength} in [0,10] to each available action. Strengths near 10 indicate actions that are in close agreement with the Advisor's rationale; strengths near 0 indicate direct opposition to it. For $n$ Advisors, $m$ actions, and comment strength $c_{ij}$ of Advisor \textit{i} on action \textit{j}, SemaFORR selects the action with the highest total comment strength: 
\[
{argmax}_{j \in m} \sum_{i=1}^n c_{ij}.
\]
\noindent 
Because ties are broken at random, tier 3 introduces uncertainty into action selection. For further details on SemaFORR, see~\cite{epstein2015learning}.

\section{Approach}
This section describes how \textsc{Why} exploits SemaFORR 
to generate natural explanations. Each of the three questions below focuses on a different aspect of a robot controller. The result is a rich, varied set of natural explanations.

\subsection{Why did you do that?}
The first question asks why the robot chose a particular action. \textsc{Why} constructs its answer from the rationales and comments of the Advisors responsible for that choice, with templates to translate actions, comments, and decisions into natural language. 
Given the robot's current pose, \textsc{Why} maps each possible action onto a descriptive phrase for use in any [action] field. Examples include ``wait'' for a forward move of 0.0 m, ``inch forward'' for a forward move of 0.2 m, and ``shift right a bit'' for a turn in place of 0.25 rad.

Algorithm \ref{alg:SEP} is pseudocode for \textsc{Why}'s responses. \textsc{Why} takes as input the current decision state, target location, and spatial model, and then calculates its response based on the comments from SemaFORR's Advisors. There are three possibilities: tier 1 chose the action, tier 1 left only one unvetoed action, or tier 3 chose the action. SemaFORR only makes a decision in tier 1 if \textsc{Victory} mandates it or \textsc{AvoidWalls} has vetoed all actions but the pause. The applicable templates in those cases are ``I could see our target and [action] would get us closer to it'' and ``I decided to wait because there's not enough room to move forward.''

The inherent uncertainty and complexity of a tier-3 decision, however, requires a more nuanced explanation. For a set of $m$ actions, assume tier-3 Advisor $D_i$ outputs comment with strengths $c_{i1},\ldots,c_{im}\in[0, 10]$. $D_i$'s \textit{t}-support for action $a_{k}$ is the \textit{t}-statistic $t_{ik} = (c_{ik} - \bar{c_i})/\sigma_i$ where $\bar{c_i}$ is the mean strength of $D_i$'s comments in the current decision state and $\sigma_i$ is their standard deviation. (This is not a $z$-score because sampled values replace the unavailable true population mean and standard deviation.) \textsc{Why} can compare different Advisors' \textit{t}-supports because they have common mean 0 and standard deviation 1. If $|t_{ik}|$ is large, Advisor $D_i$ has a strong opinion about action $a_k$ relative to the other actions: supportive for $t_{ik} > 0$ and opposed for $t_{ik} < 0$.

Table \ref{tab:example} provides a running example. It shows the original comment strengths from four Advisors on four actions, and the total strength $C_k$ for each action $a_k$. Tier 3 chooses action $a_4$ because it has maximum support. While $D_1$ and $D_2$ support $a_4$ with equal strength, the \textit{t}-support values tell a different story: $D_1$ prefers $a_4$ much more ($t_{14}=1.49$) than $D_2$ does ($t_{24}=0.71$). Moreover, $D_3$ and $D_4$ actually oppose $a_4$ ($-0.34$ and $-0.78$, respectively).

{For each measure, we partitioned the real numbers into three intervals and assigned a descriptive natural language phrase to each one, as shown in Table \ref{tab:combined}. This partitioning allows \textsc{Why} to hedge in its responses, much the way people explain their reasoning when they are uncertain~\cite{markkanen1997hedging}. \textsc{Why} maps the \textit{t}-support values into these intervals. For $a_4$, $D_1$'s \textit{t}-support of 1.49 is translated as ``want'' and $D_4$'s -0.78 is translated as ``don't want''. \textsc{Why} then completes the clause template ``I [phrase] to [rationale]'' for each Advisor based on Table \ref{tab:advisors} and less model-specific language from Table \ref{tab:combined}. For example, if $D_1$ were \textsc{Greedy}, then the completed clause template for $a_4$ would be ``I want to get close to the target.'' 

\begin{table}[!b]
\fontsize{9pt}{10pt}\selectfont
\centering
\begin{tabular}{rrrrrrrrr}
 & \multicolumn{4}{c}{$c_{ik}$}   & \multicolumn{4}{c}{$t_{ik}$} \\ \hline
 & $a_1$ & $a_2$ & $a_3$ & $a_4$ & $a_1$ & $a_2$ & $a_3$ & $a_4$\\ \hline
$D_1$ & 0 & 1 & 1 & \multicolumn{1}{r}{10} & -0.64 & -0.43 & -0.43 & 1.49 \\
$D_2$ & 0 & 8 & 9 & \multicolumn{1}{r}{10} & -1.48 & 0.27 & 0.49 & 0.71 \\
$D_3$ & 2 & 0 & 10 & \multicolumn{1}{r}{2} & -0.34 & -0.79 & 1.47 & -0.34 \\
$D_4$ & 3 & 10 & 1 & \multicolumn{1}{r}{0} & -0.11 & 1.44 & -0.55 & -0.78 \\ \hline
$C_k$ & 5 & 19 & 21 & \multicolumn{1}{r}{22} &       &       &       &      \\ 
\end{tabular}
\caption{Example of comments from tier-3 Advisors $D_i$ on actions $a_k$, where $c_{ik}$ is strength and $t_{ik}$ is \textit{t}-support}
\label{tab:example}
\end{table}

Finally, \textsc{Why} combines completed clause templates into the final tier-3 explanation, but omits language from Advisors with \textit{t}-support values in (-0.75, 0.75] because they contribute relatively little to the decision. \textsc{Why} concatenates the remaining language with appropriate punctuation and conjunctions to produce its tier-3 explanation: ``(Although [language from opposed Advisors], ) I decided to [action] because [language from supporting Advisors]''. The portion in parentheses is omitted if no opposition qualifies. If the Advisors in the running example were \textsc{Greedy}, \textsc{ElbowRoom}, \textsc{Convey}, and \textsc{Explorer}, in that order, and $a_4$ were move forward 1.6 m, then the natural explanation is ``Although I don't want to go somewhere I've been, I decided to move forward a lot because I want to get close to our target.'' (Note that $D_2$'s support fails the -0.75 filter and so is excluded.)

This approach can also respond to ``What action would you take if you were in another context?'' Given the decision state and the target location, \textsc{Why} would reuse its current spatial model, generate hypothetical comments, and process them in the same way. The sentence template would substitute ``I would [action]'' for ``I decided to [action]." 

\subsection{How sure are you that this is the right decision?}
The second question from a human collaborator is about the robot's confidence in its decision, that is, how much it trusts that its decision will help reach the target. Again, \textsc{Why} responds based on the tier that selected the action. Tier 1's rule-based choices are by definition highly confident. If \textsc{Victory} chose the action then the response is ``Highly confident, since our target is in sensor range and this would get us closer to it.'' If \textsc{AvoidWalls} vetoed all forward moves except the pause, then the explanation is ``Highly confident, since there is not enough room to move forward.'' 

Again, tier-3's uncertainty and complexity require more nuanced language, this time with two measures: level of agreement and overall support. The extent to which the tier-3 Advisors agree indicates how strongly the robot would like to take the action. \textsc{Why} measures the level of that agreement with Gini impurity, where values near 0 indicate a high level of agreement and values near 0.5 indicate disagreement~\cite{hastie2009elements}. For $n$ tier-3 Advisors and maximum comment strength 10, the \textit{level of agreement} $G_k$ $\in$ [0,0.5] on action $a_k$ is defined as
\[
G_k = 2 \cdot \left[\frac{\sum_{i=1}^n c_{ik}}{10n}\right] \cdot \left[1-\frac{\sum_{i=1}^n c_{ik}}{10n}\right].
\]
 In the example of Table \ref{tab:example}, the level of agreement on $a_4$ is 
$G_4 = 2 \cdot \left[\frac{22}{40}\right] \cdot \left[1-\frac{22}{40}\right] \approx 0.50$. This indicates considerable disagreement among the Advisors in Table \ref{tab:example}.

The second confidence measure is SemaFORR's overall support for its chosen action compared to other possibilities, defined as a $t$-statistic across all tier-3 comments. Let $\mu_C$ be the mean total strength of all actions \textit{C} under consideration by tier 3, and $\sigma_C$ be their standard deviation. We define the \textit{overall support} for action $a_k$ as $T_k = (C_{k} - \mu_C)/\sigma_C$. $T_k$ indicates how much more the Advisors as a group would like to perform $a_k$ than the other actions. In Table \ref{tab:example}, the overall support $T_4$ for $a_4$ is 0.66, which indicates only some support for $a_4$ over the other actions.
\begin{table}[!b]
    \fontsize{9pt}{10pt}\selectfont
    \centering
\begin{tabular}{p{0.75 in} p{0.72 in} p{1.31 in}}
\hline
\multirow{3}{\hsize}{t-support $t_{ik} \leq 0$: opposed} & $(-\infty, -1.5]$ & really don't want \\ 
                  & $(-1.5, -0.75]$ & don't want \\ 
                  & $(-0.75, 0]$ & somewhat don't want \\ \hline
\multirow{3}{\hsize}{t-support $t_{ik} > 0$: supportive} & $(0, 0.75]$ & somewhat want \\ 
                  & $(0.75, 1.5]$ & want \\ 
                  & $(1.5, +\infty)$ & really want \\ \hline
\multirow{3}{\hsize}{Level of agreement $G_k$} & $(0.45, 0.5]$ & My reasons conflict \\ 
                  & $(0.25, 0.45]$ & I've only got a few reasons \\ 
                  & $[0, 0.25]$ & I've got many reasons \\ \hline
\multirow{3}{\hsize}{Overall support $T_k$} & $(-\infty, 0.75]$ & don't really want \\ 
                  & $(0.75, 1.5]$ & somewhat want \\ 
                  & $(1.5, +\infty)$ & really want \\ \hline
\multirow{3}{\hsize}{Confidence level $L_k$} & $(-\infty, 0.0375]$ & not \\ 
                  & $(0.0375, 0.375]$ & only somewhat \\ 
                  & $(0.375, +\infty)$ & really \\ \hline
\multirow{3}{\hsize}{Difference in overall support $T_k - T_j$} & $(0, 0.75]$ & slightly more \\ 
                  & $(0.75, 1.5]$ & more \\
                  & $(1.5, +\infty)$ & much more \\ \hline
\end{tabular}
    \caption{Phrase mappings from value intervals to language}
    \label{tab:combined}
\end{table}

\textsc{Why} weights level of agreement and overall support equally to gauge the robot's confidence in a tier-3 decision with \textit{confidence level} $L_k = (0.5 - G_k) \cdot T_k$ for $a_k$. It then maps each of $L_k$, $G_k$, and $T_k$ to one of three intervals and then to natural language, as in Table \ref{tab:combined}, with implicit labels \textit{low} $< $ \textit{medium} $<$ \textit{high} in order for each statistic. Two statistics \textit{agree} if they have the same label; one statistic is lower than the other if its label precedes the other's in the ordering.

All responses to this question use a template that begins ``I'm [$L_k$ adverb] sure because....'' If $G_k$ and $T_k$ both agree with $L_k$, the template continues ``[$G_k$ phrase]. [$T_k$ phrase].'' For example, ``I'm really sure about my decision because I've got many reasons for it. I really want to do this the most.'' If only one agrees with $L_k$, the template continues ``[phrase for whichever of $G_k$ or $T_k$ agrees].'' For example, ``I'm not sure about my decision because my reasons conflict.'' Finally, if neither agrees with $L_k$, it concludes ``even though [phrase for whichever of $G_k$ or $T_k$ is lower than $L_k$], [$G_k$ phrase or $T_k$ phrase that is higher than $L_k$].'' For example, ``I am only somewhat sure about my decision because, even though I've got many reasons, I don't really want to do this the most.'' For $a_4$ in Table \ref{tab:example}, $L_4$ is near 0, $G_4$ = 0.50, and $T_4$ = 0.66. This produces the natural explanation ``I'm not sure about my decision because my reasons conflict. I don't really want to do this more than anything else.''

\begin{figure}[!b]
    \centering
    \includegraphics[width = 0.7\linewidth]{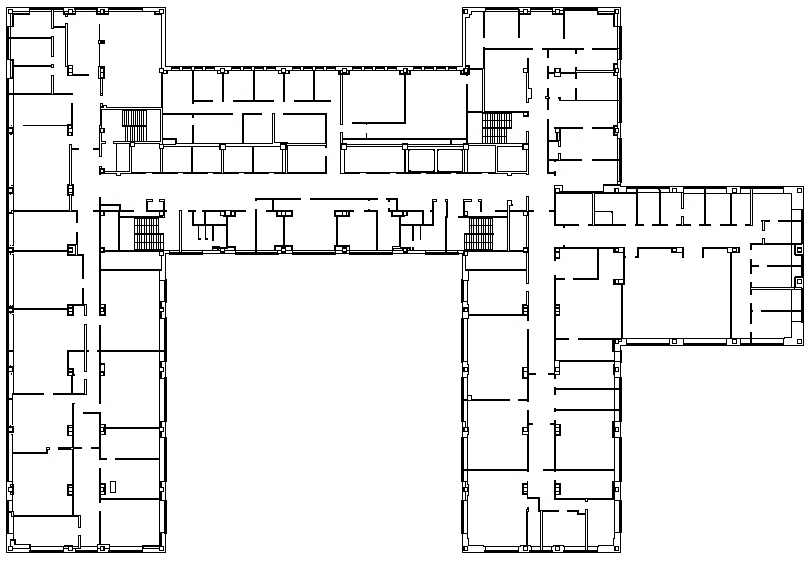}
    \caption{Tenth floor of Hunter College's North Building}
    \label{fig:resultsexamples}
\end{figure}

\subsection{Why not do something else?}
A human collaborator makes decisions with her own mental model of the world. When her decision conflicts with another team member's, she tries to understand why they made a different decision. \textsc{Why}'s approach explains SemaFORR's preference for action $a_k$ over an alternative $a_j$. If tier 1 chose $a_k$, the explanation uses \textsc{Victory}'s rationale: ``I decided not to [action$_j$] because I sense our goal and another action would get us closer to it.'' If \textsc{AvoidWalls} or \textsc{NotOpposite} vetoed $a_j$, then the natural explanation is ``I decided not to [action] because [rationale from Advisor that vetoed it].'' 

The other possibility is that $a_j$ had lower total strength in tier 3 than $a_k$ did. In this case, \textsc{Why} generates a natural explanation with the tier-3 Advisors that, by their comment strengths, discriminated most between the two actions. \textsc{Why} calculates $t_{ik}-t_{ij}$ for each Advisor $D_i$. If the result lies in [-1, 1] then $D_i$'s support is similar for $a_k$ and $a_j$; otherwise $D_i$ displays a \textit{clear preference}. The natural explanation includes only those Advisors with clear preferences.

The explanation template is ``I thought about [action$_j$] (because it would let us [rationales from Advisors that prefer action$_j$]), but I felt [phrase] strongly about [action$_k$] since it lets us [rationales from Advisors that prefer action$_k$].'' The [phrase] is the extent to which SemaFORR prefers $a_k$ to $a_j$. It is selected based on $T_k-T_j$, the difference in the actions' overall support, and mapped into intervals as in Table \ref{tab:combined}. The portion in parentheses is only included if any Advisors showed a clear preference for action$_j$.

For ``Why didn't you take action $a_2$?'' on our running example, \textsc{Why} calculates the difference in overall support between $a_4$ and $a_2$ at 0.38, which maps to ``slightly more.'' The differences in $t$-support between $a_4$ and $a_2$ are 1.92, 0.44, 0.45, and -2.22. Thus, if $D_1$ is \textsc{Greedy} and prefers $a_4$, while $D_4$ is \textsc{Explorer} and prefers $a_2$, the natural explanation is ``I thought about $a_2$ because it would let us go somewhere new, but I felt slightly more strongly about $a_4$ since it lets us get closer to our target.''

\begin{table}[!b]
    \fontsize{9pt}{10pt}\selectfont
    \centering
    \begin{tabular}{p{1.48 in} rrr}
        \hline
        Tier where made & 1 & 3 & All\\
        \hline
        Number of decisions & 22,982 & 84,920 & 107,902\\
        Avg. computation time (ms) & 0.45 & 3.08 & 2.52\\
       \hline
       Unique phrasings  \\ \hline
        Why? & 14 & 31,896 & 31,910\\
        Confidence? & 2 & 11 & 13\\
        Something else? & 19 & 124,086 & 124,105\\
        Total & 35 & 155,993 & 156,028\\
        \hline
       Average readability  \\ \hline
        Why? & 8.18 & 5.02 & 5.70\\
        Confidence? & 10.39 & 7.63 & 8.22\\
        Something else? & 3.91 & 6.44 & 5.96\\
        Overall & 5.36 & 6.41 & 6.21\\
    \end{tabular}
    \caption{Empirical explanations}
    \label{tab:results}
\end{table}

\begin{table}[!b]
    \fontsize{9pt}{10pt}\selectfont
    \centering
\begin{tabular}{| l | r | r | r |}
\hline
 & \multicolumn{1}{c|}{Low} & \multicolumn{1}{c|}{Medium}  & \multicolumn{1}{c|}{High} \\ \hline
$G_k$ & 67.15\%& 30.41\%& 2.44\%                     \\  
$T_k$ & 2.34\%& 60.09\%& 37.57\%                     \\  
$L_k$ &  54.92\%& 42.64\%& 2.44\%                      \\  
$t_{ik} - t_{ij}$ &  16.09\%& 44.41\%& 39.50\%                      \\ 
$T_k - T_j$ &  18.48\%& 20.40\%& 61.13\%                    \\ 
\hline
\end{tabular}
    \caption{Metric distributions by interval in tier-3 decisions}
    \label{tab:histograms}
\end{table}

\begin{table*}[!t]
    \fontsize{9pt}{10pt}\selectfont
    \centering
\begin{tabular}{| p{1.56 in} | p{1.56 in} | p{1.56 in} | p{1.56 in} |}
\hline
\multicolumn{4}{|c|}{Decision State} \\ 
\hline
\raisebox{-.5\height}{\includegraphics[width = \linewidth]{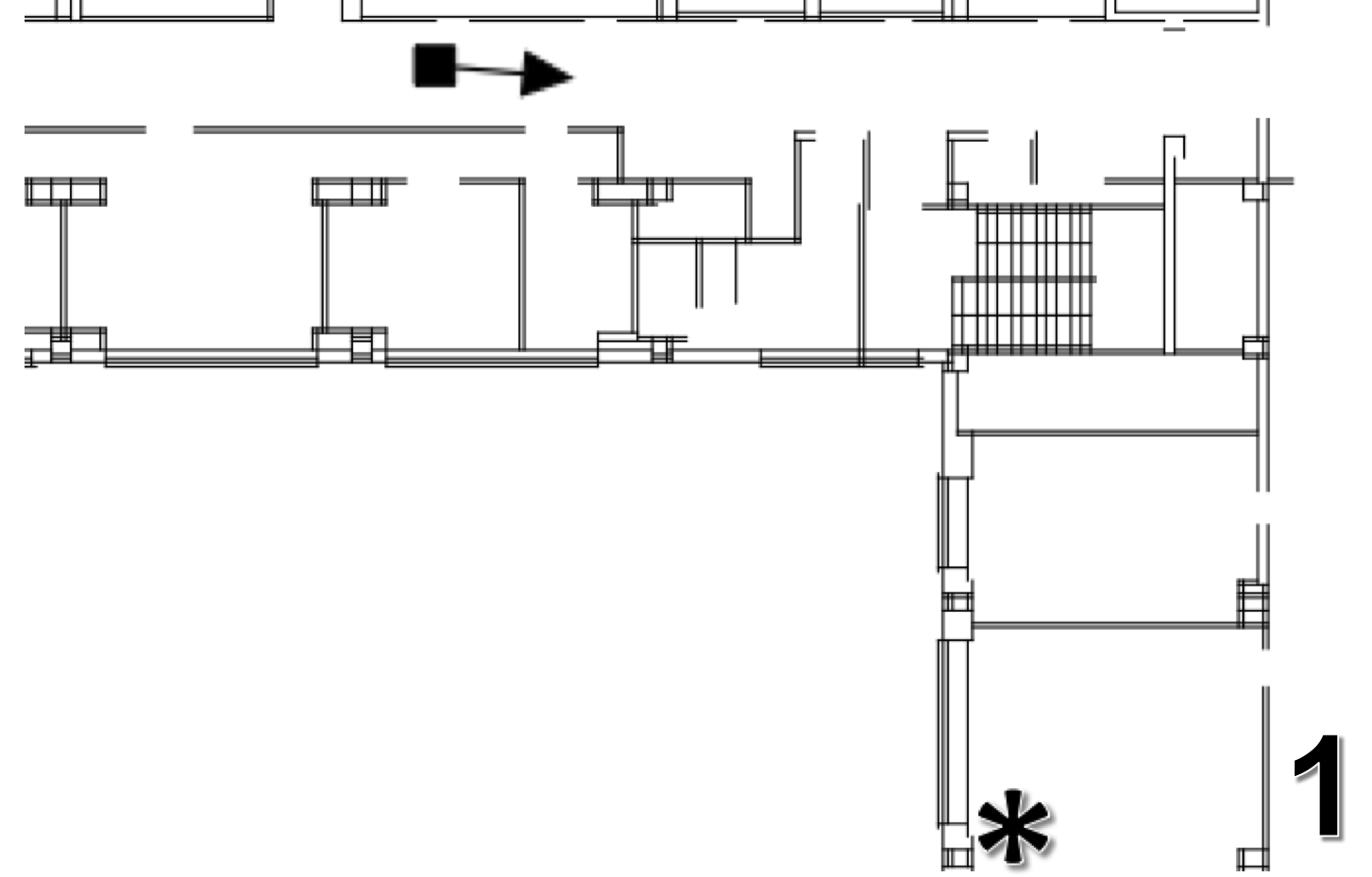}} & \raisebox{-.5\height}{\includegraphics[width = \linewidth]{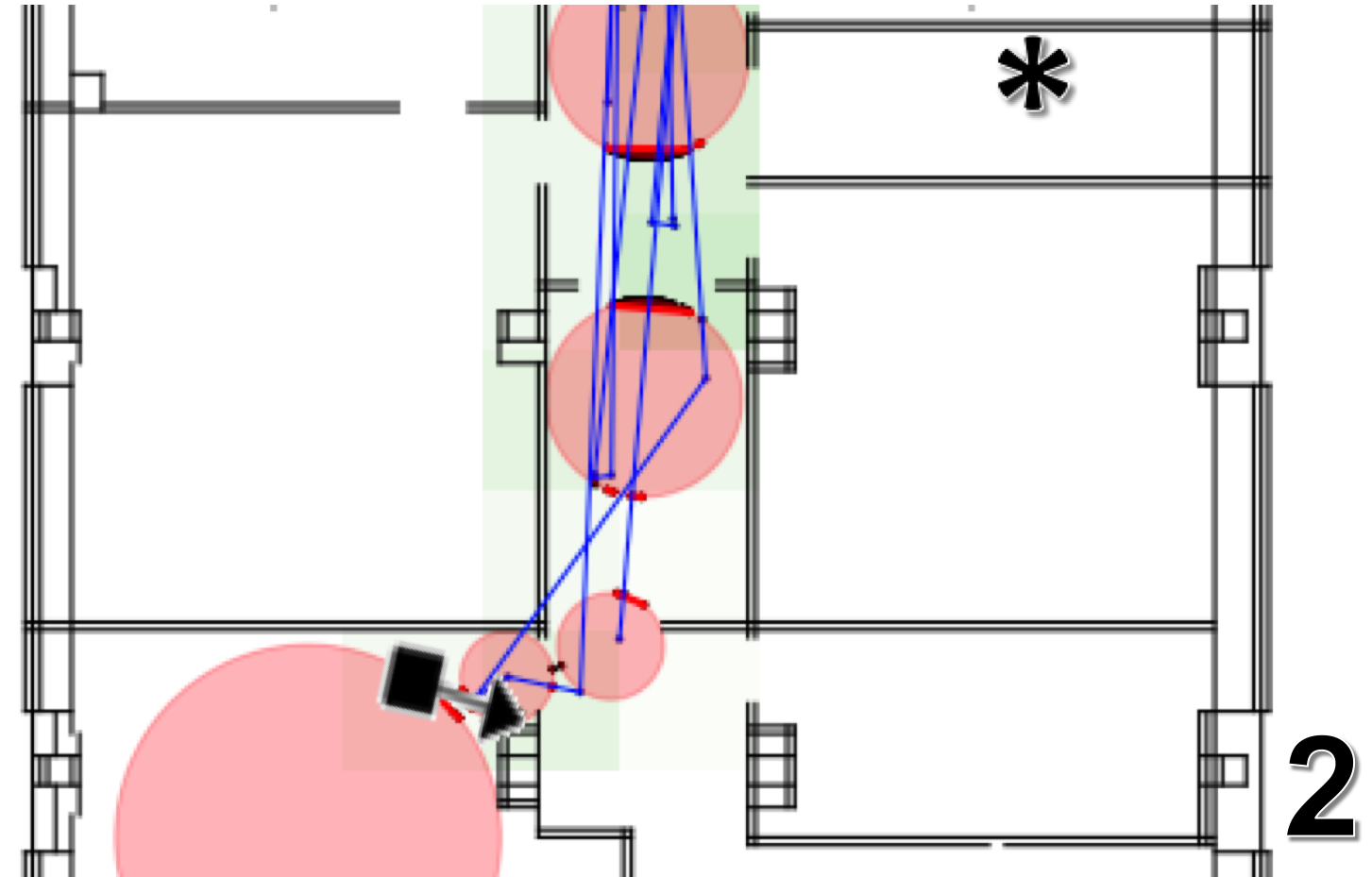}} & \raisebox{-.5\height}{\includegraphics[width = \linewidth]{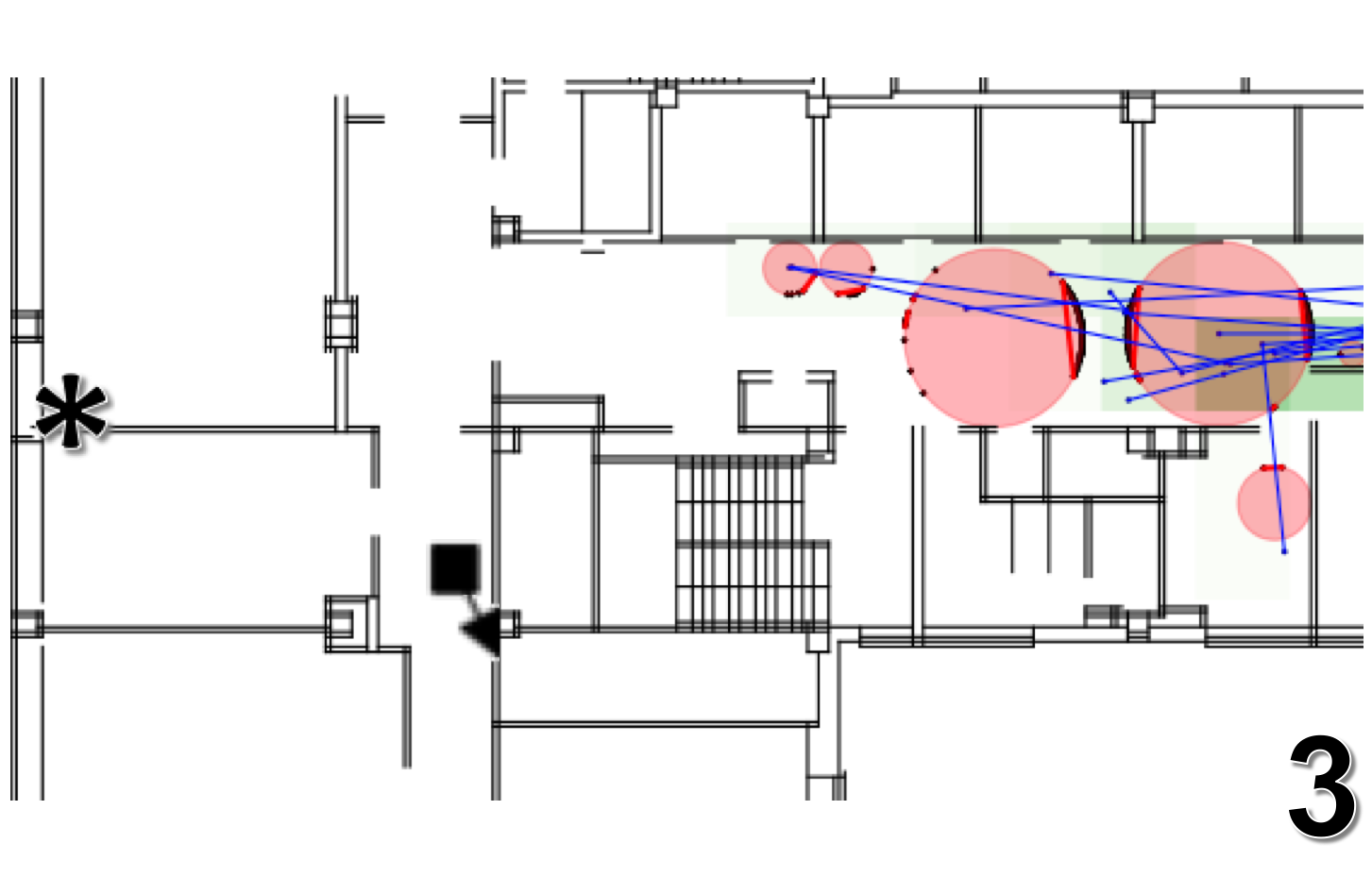}} & \raisebox{-.5\height}{\includegraphics[width = \linewidth]{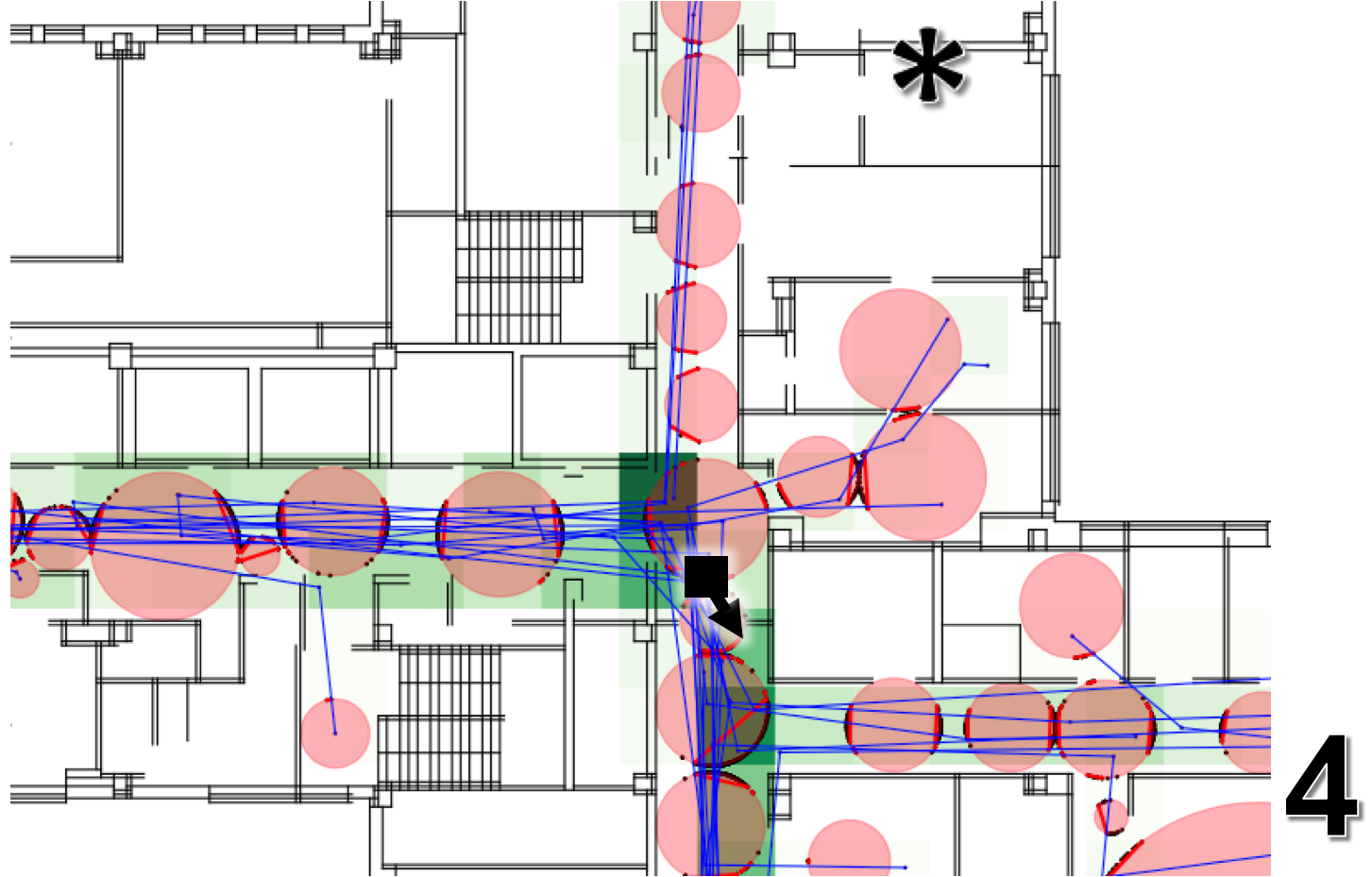}} \\ \hline
\multicolumn{4}{|c|}{Why did you do that?} \\ \hline
Although I don't want to go close to that wall, I decided to bear right because I really want to take a big step. & Although I don't want to turn towards this wall, I decided to turn right because I want to go somewhere familiar, I want to get close to our target, and I want to follow a familiar route that gets me closer to our target. & Although I really don't want to go close to that wall and I really don't want to get farther from our target, I decided to move forward a lot because I really want to go to an area I've been to a lot, I really want to take a big step, and I really want to go somewhere new. & Although I don't want to get farther from our target, I decided to bear left because I really want to go somewhere familiar and I want to leave since our target isn't here. \\ \hline
\multicolumn{4}{|c|}{How sure are you?} \\ \hline
I'm not sure because my reasons conflict. & I'm only somewhat sure because, even though my reasons conflict, I really want to do this most. & I'm not sure because my reasons conflict. & I'm only somewhat sure in my decision because I've only got a few reasons. I somewhat want to do this most. \\ \hline
\multicolumn{4}{|c|}{Why not do something else?} \\ \hline
I thought about turning left because it would let us stay away from that wall and get close to our target, but I felt more strongly about bearing right since it lets us take a big step and get around this wall. & I thought about shifting left a bit because it would let us get around this wall, but I felt much more strongly about turning right since it lets us go somewhere familiar and get close to our target. & I decided not to move far forward because the wall was in the way. & I thought about turning hard right because it would let us get close to our target, but I felt much more strongly about bearing left since it lets us go somewhere familiar, leave since our target isn't here, go somewhere new, and get around this wall. \\ \hline
\end{tabular}
    \caption{Explanations for decision states and any current spatial model, enlarged from Figure \ref{fig:resultsexamples}}
    \label{tab:resultsexplanations}
\end{table*}

\section{Results}
Implemented as a ROS package, \textsc{Why} explains Sema-FORR's decisions in real time. We evaluated \textsc{Why} in simulation for a real-world robot (Fetch Robotics' Freight). When the robot navigated to 230 destinations in the complex 60m$\times$90m office world of Figure \ref{fig:resultsexamples}, \textsc{Why} averaged less than 3 msec per explanation.

\textsc{Why}'s many distinct natural explanations simulate people's ability to vary their explanations based on their context~\cite{malle1999people}. Table \ref{tab:results} provides further details. The Coleman-Liau index measures text readability; it gauged \textsc{Why}'s explanations over all three questions at approximately a 6th-grade level~\cite{coleman1975computer}, which should make them readily understandable to a layperson. 

For action $a_k$ chosen in tier 3 and every possible alternative $a_j$, Table \ref{tab:histograms} shows how often the values of $G_k$, $T_k$, $L_k$, $t_{ik}-t_{ij}$, and $T_k-T_j$ fell in their respective Table \ref{tab:combined} intervals. The Advisors disagreed ($G_k > 0.45$) on 67.15\% of decisions. Strong overall support ($T_k > 1.5$) made SemaFORR strongly confident in 2.44\% of its decisions ($L_k > 0.375$) and somewhat confident in 42.64\% of them. When asked about an alternative, individual Advisors clearly preferred ($t_k-t_j > 1$) the original decision 39.50\% of the time; SemaFORR itself declared a strong preference ($T_k-T_j > 1.5$) between the two actions 61.13\% of the time.

Table \ref{tab:resultsexplanations} illustrates \textsc{Why}'s robust ability to provide nuanced explanations for tier-3 decisions. The target appears as an asterisk and the black box and arrow show the robot's pose. Decision 1 was made when the robot had not yet learned any spatial affordances; decision 2 was made later, when the spatial model was more mature. In decision 3, the Advisors strongly disagreed, while in decision 4 the spatial model-based Advisors disagreed with a commonsense-based Advisor. 

\section{Discussion}
\textsc{Why} is applicable more broadly than we have indicated thus far. Any robot controller could have SemaFORR learn the spatial model in parallel, and use it with \textsc{Why} to produce transparent, cognitively-plausible explanations. If the alternative controller were to select action $a_j$ when SemaFORR selected $a_k$, \textsc{Why} could still explain $a_j$ with any Advisors that supported it, and offer an explanation for $a_k$ as well. Furthermore, once equipped with Advisor phrases and possibly with new mappings, any FORR-based system could use \textsc{Why} to produce explanations. For example, Hoyle is a FORR-based system that learns to play many two-person finite-board games expertly~\cite{epstein2001learning}. For Hoyle, \textsc{Why} could explain ``Although I don't want to make a move that once led to a loss, I decided to do it because I really want to get closer to winning and I want to do something I've seen an expert do.''

Because SemaFORR's spatial model is approximate and its Advisors are heuristic, precise natural language interpretations for numeric values are ad hoc. For Table \ref{tab:combined}, we inspected thousands of decisions, and then partitioned the computed values as appeared appropriate. We intend to fine-tune both intervals and phrasing with empirical assessment by human subjects. Because natural explanations have improved people's trust and understanding of other automated systems, we will then evaluate \textsc{Why} with human subjects. 

SemaFORR and \textsc{Why} are both ongoing work. As heuristic planners for tier 2 are developed, we will extend \textsc{Why} to incorporate plans in its explanations. We also anticipate revisions in \textsc{Why}'s phrasing to reflect changes in SemaFORR's possible action set. Finally, \textsc{Why} could be incorporated into a more general dialogue system that would facilitate part of a broader conversation between a human collaborator and a robot. A FORR-based system for human-computer dialogue, could prove helpful there~\cite{epstein2011role}.

In summary, \textsc{Why} produces natural explanations for a robot's navigation decisions as it travels through a complex world. These explanations are essential for collaborative navigation and are made possible by the robot controller's cognitively-based reasoning. The approach presented here generates explanations that gauge the robot's confidence and give reasons to take an action or to prefer one action over another. As a result, a human companion receives informative, user-friendly explanations from a robot as they travel together through a large, complex world in real time.

\noindent \textbf{Acknowledgements.} This work was supported in part by NSF 1625843. The authors thank the reviewers for their insightful comments.

\fontsize{9pt}{10pt}\selectfont
\bibliographystyle{aaai}
\bibliography{bibliography}

\end{document}